\documentclass[final,5p,times,twocolumn]{elsarticle}

\usepackage{amssymb}
\usepackage{amsthm}

\usepackage{array}
\usepackage{slashbox}
\usepackage{mdwmath}
\usepackage{mdwtab}
\usepackage{amsmath}
\usepackage{url} 

\journal{Pattern Recognition}

\begin{document}

\begin{frontmatter}

\title{Cross-pose Face Recognition by Canonical Correlation Analysis}

 \author[label1]{Annan Li}
  \author[label2]{Shiguang Shan}
   \author[label2]{Xilin Chen}
   \author[label3]{Bingpeng Ma}
   \author[label1]{Shuicheng Yan}
    \author[label4]{Wen Gao}

\address[label1]{Department of Electrical and Computer Engineering, National University of Singapore, Singapore. }
 \address[label2]{Institute of Computing Technology, Chinese Academy of Sciences, Beijing,China}
 \address[label3]{University of Chinese Academy of Sciences, Beijing, China}
\address[label4]{Institute of Digital Media, Peking University, Beijing, China}


\begin{abstract}
The pose problem is one of the bottlenecks in automatic face recognition. We argue that one of the difficulties in this problem is the severe misalignment in face images or feature vectors with different poses. In this paper, we propose that this problem can be statistically solved or at least mitigated by maximizing the intra-subject across-pose correlations via canonical correlation analysis (CCA). In our method, based on the data set with coupled face images of the same identities and across two different poses, CCA learns simultaneously two linear transforms, each for one pose. In the transformed subspace, the intra-subject correlations between the different poses are maximized, which implies pose-invariance or pose-robustness is achieved. The experimental results show that our approach could considerably improve the recognition performance. And if further enhanced with holistic+local feature representation, the performance could be comparable to the state-of-the-art.
\end{abstract}


\end{frontmatter}


\section{Introduction}
\label{sec_intro}
Automatic face recognition is a classical research topic in computer vision and pattern recognition research. After more than 30 years of research, the performances of face recognition systems have been greatly improved. In some evaluation tests computer based face recognition systems even outperform humans \cite{pami07surpasshumans}. However, the high performance is usually achieved under controlled imaging conditions. Usually, it means that the face images are acquired under frontal view,  normal expression and mild illumination. These requirements are often unrealistic in real-world applications, since the variations of pose, illumination and expression are very common and uncontrollable. When these variations are present, the performance of face recognition often drops significantly \cite{face03survey}. Thus, the problem of face recognition is far from being solved. In above-mentioned challenging problems, the pose problem is one of the most important and difficult issues for face recognition. In this paper we present a novel pose robust face recognition approach that can significantly improve the recognition performance.
\begin{figure}[ht]
\begin{center}
   \includegraphics[width=0.52\linewidth]{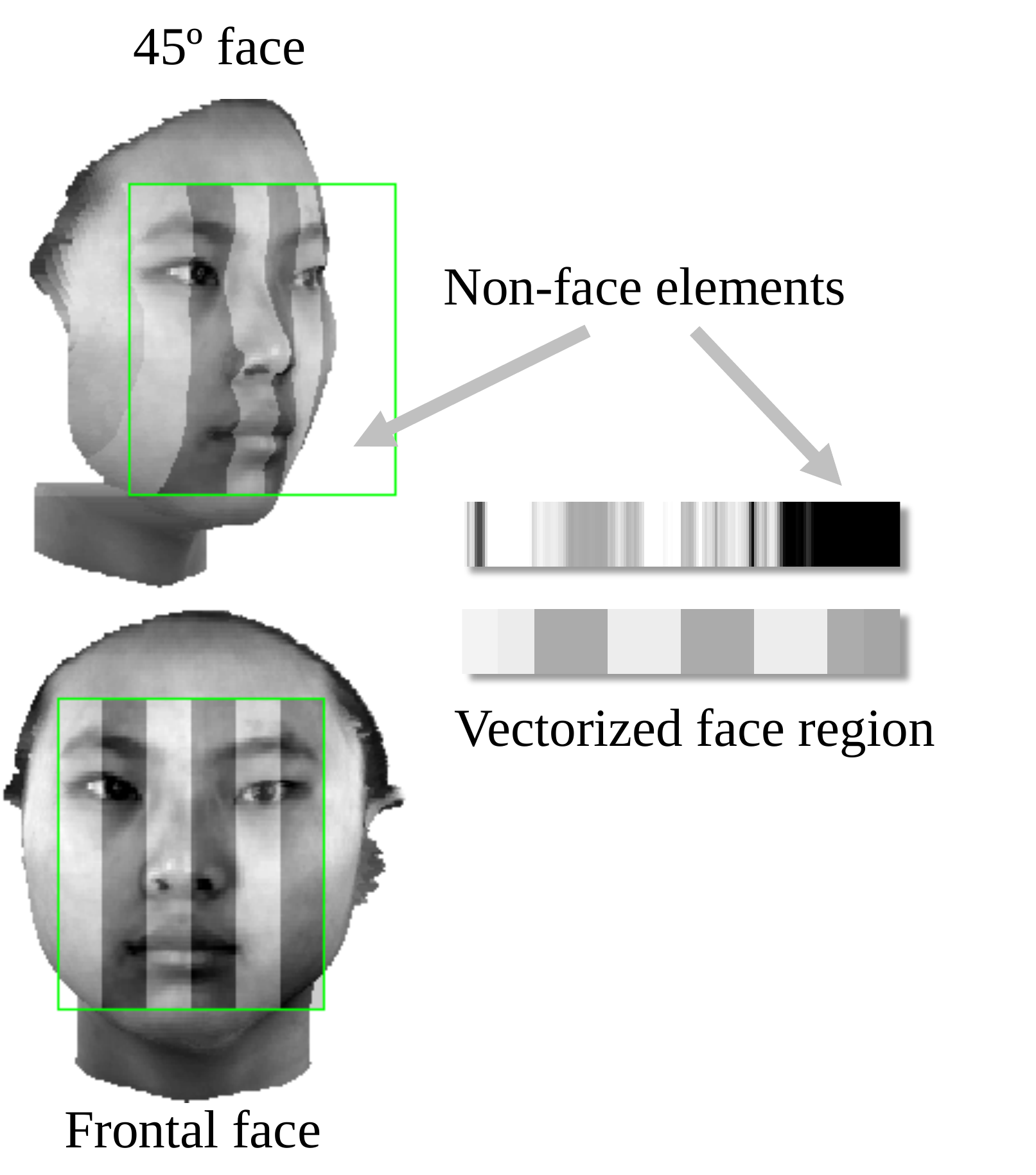}
   \vspace{-3mm}
   \caption{Pose variations lead to misalignment and noise in the feature vectors. The face images under frontal view and $45^{\circ}$ are generated by a 3D face model. The corresponding face regions are given by the ground truth 3D shape and illustrated in the same color. The feature vectors are misaligned. And the feature vector of non-frontal face even contains non-face elements.}
   \label{fig:pose_misalign}
   \end{center}
\end{figure}

\begin{figure}[h]
\begin{center}
   \includegraphics[width=0.6\linewidth]{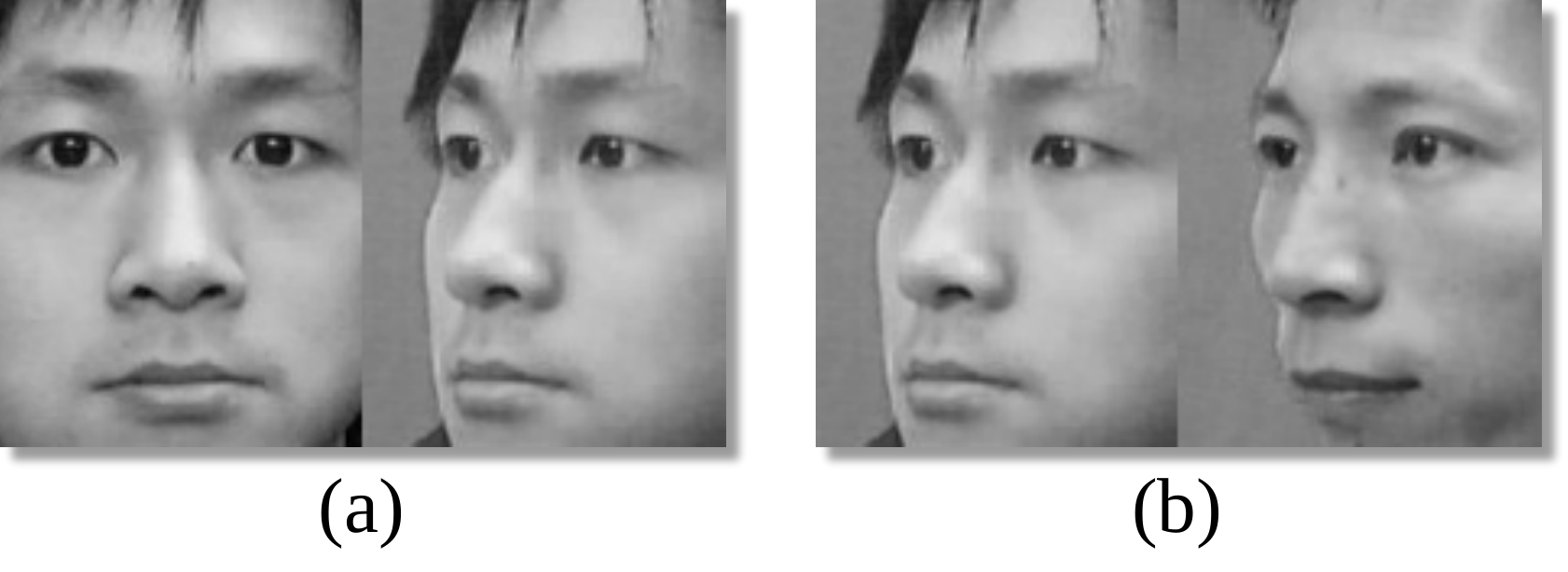}
   \vspace{-3mm}
   \caption{In vector based approaches, faces in similar pose is more similar than those in different pose even for the faces of the same identity.}
   \label{fig:pose_problem}
   \end{center}
\end{figure}

The variations of appearance caused by pose differences in 2D face images are related to two factors, i.e., the viewpoint and the 3D facial shape. The 3D shape of human face has a complex structure. Thus, in the 2D image taken from a different viewpoint, the locations of surface points on face change differently. That is to say, a pair of points close in one pose may become far away from each other in another pose. This inner structure distortion in image makes the alignment very difficult. Furthermore, the 3D shape of human face is not ideally convex. Its concavity area leads to occlusion. That is, when pose changes, some visible parts on face may become invisible while some invisible parts may become visible. In the vector based approaches, the elements in the feature vector are sampled continuously with equal step on the 2D face image or on the output of some filters on the face image, such as the Gabor filters \cite{TIP02gabor_liu}. However, when pose difference is big the inner distortion and occlusion mentioned above lead to misalignment and noise in the feature vectors. As shown in Figure 1, if the elements are sampled in the same way on frontal and non-frontal faces, the feature vectors are misaligned, and even contain some non-face elements. It is not surprising that the performance of face recognition would be poor using such feature vectors. It also explains a special phenomenon in vector based cross-pose face recognition that the distance between two faces of different people under similar viewpoint is often smaller than that of the same person under different viewpoints (see Figure \ref{fig:pose_problem}). Viewed from this point, the key problem for pose-robust face recognition is how to measure the similarity of identity between two misaligned and noisy feature vectors.

Aligning the elements in feature vectors is equal to reconstruct the fine 3D shape of face. Because of the complexity of 3D face, misalignment is a very challenging problem. However, it does not mean that the problem is insoluble. Although the human face varies from person to person, the variations caused by pose difference in 2D face images have particular statistical character in common. In a specific pose, the faces of a given person can be well represented by a linear subspace. The difficulty of the pose problem is that the subspace of different pose is different due to the misalignment mentioned above. However, these different subspaces can be aligned using the face pairs coupled by identity \cite{pie_db_pami,cas_peal_db}, which is shown in Figure 3. So, the problem of recognizing faces across poses can be formulated as how to measure the similarity between two vectors from different subspaces. We proposed that this theoretical problem can be elegantly solved by using the \emph{Canonical Correlation Analysis} (CCA) method \cite{hotelling1936cca}. The CCA can maximize the correlations between two different sets of variables. Thus, if performing CCA on the face pairs coupled by identity across two poses, the intra-subject correlations between two poses can be maximized. Consequently the problem caused by misalignment is statistically avoided or mitigated. 

Based on the above analysis, we propose a novel approach for recognizing face across poses. In this approach, two sets of pose specific basis vectors are learned simultaneously based on the coupled face data. Feature vectors from different poses are projected onto the basis vectors respectively, and then face recognition across pose differences is performed by matching these projections. The basic idea of this approach is illustrated in Figure 3. The CCA was first applied to tackle the pose problem in our preliminary work \cite{annan_cvpr09}, in which it is mainly used as a patch matching and virtual view synthesis method. Different from it, in this paper, we extend the proposed CCA based recognition approach by adopting facial feature representation method that integrates holistic features and local Gabor features. Based on this feature representation method, multiple classifiers are built via CCA and combined together to enhance the recognition performance.

The remaining parts of this paper are organized as follow. Section \ref{sec_related_works} gives a brief review of cross-pose face recognition, and Section \ref{sec_cca} describes our method in detail. Subsequently, Section \ref{sec_exp} presents the experimental results and lastly, we draw the conclusions and discuss the future works in Section \ref{sec_conclusions}.

\section{Related Works}
\label{sec_related_works}

Robustness to pose change is a challenging and classical problem in the research of face recognition. Related literature survey can be found in \cite{PR09PoseSurvey,lucey2008viewpoint,face03survey}. When multiple face images under different view are available, the difficulty of the pose problem is much reduced. But in real world applications such requirement is not always feasible. In this paper, we only concern the basic and most challenging scenario: recognizing face across pose difference using single query image. According to the main contribution, related research works can be roughly categorized into three classes, i.e., the geometrical approaches, the statistical approaches and the hybrid approaches respectively.

As described in previous section, misalignment is the key factor that leads to performance degradation when pose variations are present. So, pursuing geometric alignment of face is a natural way to tackle the pose problem. Beymer and Poggio \cite{beymer_iccv1995} used single example 2D image to predict non-frontal faces from frontal faces. In this approach, dense pixel to pixel alignment is obtained from a pair of 2D example images. In some sense, pursuing the precise pixel to pixel alignment across different viewpoint is equivalent to reconstructing the 3D shape of face. When 3D face data is available, many approaches of recovering 3D facial shape from 2D image are proposed. Among them, the 3D morphable model proposed by Blanz and Vetter \cite{pami03_3dmm} is considered the state of the art. By fitting the statistical 3D model to the input face, high recognition rate can be achieved using the representation coefficients or the transformed images \cite{cvpr05Blanz}. Although the optimization process of fitting guarantees the reconstruction accuracy, it brings the problem of high computational complexity. An extension of this work with spherical harmonic illumination model can be found in \cite{pami06_3dharmonic}. Recently, Asthana et al. \cite{iccv11full_auto} use active appearance models (AAMs) \cite{pami01aam} to detect facial landmarks and reconstruct 3D facial shape based on the landmarks. Cross-pose face recognition is performed by matching synthesized face using local binary patterns\cite{pami06lbp}. Mostafa et al. \cite{eccv12at_distance} proposed a similar work in which active shape models (ASMs) \cite{cootes1995asm} are utilized for facial landmark detection.

Although methods like 3D morphable model achieves good performance, accurate 3D face reconstruction is still a complicated and difficult problem. Prabhu et al. \cite{pami11_3d_gem} argued that depth information of faces may be not significantly discriminative for modeling 2D pose variability. Thus, if one could obtain several such generic canonical depth maps for different input face groups, such as race, age, and gender, then face images under different poses can be easily rendered. Consequently, they proposed a 3D Generic Elastic Models for generating new views of template face. Face recognition across pose is performed by matching the input faces with the rendered faces. Besides the work of \cite{pami11_3d_gem}, Castillo and Jacobs \cite{cvpr07stereo_matching,pami09stereo_matching} also proposed a simplified geometrical approach for cross-pose face recognition. They simplified the 3D shape of face as a cylinder and recognized face through stereo matching. To address the problem of large pose difference, an improvement with surface slant of this approach was proposed recently \cite{cvpr2011wide_bsl_stereo}. Since it does not need to perform 3D reconstruction, recognizing face across pose via stereo matching is simple and effective. 

Different from the dense point-to-point alignment approaches mentioned above, sparse alignment approaches only use several facial landmarks, such as the eye corners and nose tips etc. These landmarks are usually salient in appearance. Thus, the difficulty of alignment and the computational cost could be greatly reduced. Wiskott et al. \cite{pami97EBGM} proposed the Elastic Bunch Graph Matching method, in which several fiducial points are elastically matched using local Gabor feature. The AAMs \cite{pami01aam} is similar to the 3D morphable model. A major difference between them is the shape model: in AAMs, the morphable model is simplified to some facial landmarks in 2D space. Some multi-view extension of AAMs can be found in \cite{bmvc00aam,cootes02view_based_aam,fg02_stan_aam}.

Besides the geometrical approaches, building statistical models is another popular way to recognize faces across poses. Hitherto, a typical statistical approach is the eigen light-field method proposed by Gross et al. \cite{PAMI04ELF}. They built a complete appearance model including all possible pose variations. A test image can be viewed as a part of this complete model. The missing parts are estimated from the available parts. The recognition is performed by comparing the coefficients of the complete appearance model. The tied factor analysis method proposed by Prince et al. \cite{pami08tiedfactor} is another typical statistical approach. In this method some tied factors across pose difference are learned using Expectation Maximization algorithm. Then, face recognition is performed based on the  probabilistic distance metric that built on the factor loadings. Besides principal component analysis and factor analysis, Li et al. \cite{prl2011annan} Sharma and Jacobs \cite{cvpr11pls} applied partial least squares in cross-pose face recognition.  They used partial least squares to learn a pair of projection matrix for two different poses, and cross-pose face recognition is performed by comparing the ``intermediate correlated projections''.

Besides building the pose-robust statistical models, statistically transforming face or features from one pose to another is another way to tackle the pose problem. Sanderson et al. \cite{PR06sanderson} transformed the frontal face model to non-frontal views for extending the gallery set. Lee and Kim \cite{Kim_prl06} constructed feature spaces for each pose using kernel principal component analysis, and then transformed the non-frontal face to frontal through the feature spaces. Different from the foregoing two methods that transform holistic faces, Chai et al. \cite{TIP07LLR} performed linear regression on local patches for virtual frontal view synthesis. Li et al. \cite{tip2012annan} embedded bias-variance trade-off in the cross-pose linear regression models by using ridge and lasso regression. Such bias-variance trade-off achieved considerable improvements in recognition performance. Choi et al. \cite{prl11pose_illum} applied null space linear discriminant analysis in their face recognition approach dealing with both pose and illumination variations.

The geometrical alignment can directly reduce the pose difference, but the alignment itself cannot deal with the occlusion. Statistical approaches can mitigate the occlusion problem to some extent, their performance highly relies on the training data. Thus, combining geometrical and statistical information is an alternative way to tackle the pose problem. We term this kind of methods the hybrid approach. One way to integrate geometrical and statistical information is combining local statistical models with coarse geometrical alignment. Kanade and Yamada \cite{KanadeYamada03} proposed a probabilistic framework to build and combined local statistical models for recognizing faces under different viewpoints, in which the statistical models are built on local patches on face images. Based on this framework, Liu and Chen \cite{cvpr05liuxiaoming} used a simple 3D ellipsoid model to align patches across different pose, while Ashraf et al. \cite{cvpr08Ashraf} aligned the patches by learning 2D affine transform for each patch pair via a Lucas-Kanade \cite{LK1981Darpa} like optimization procedure.  Unlike the above methods that concern on matching local patches, Lucey and Chen \cite{cvpr06Lucey} extended Kanade and Yamada's work by building similar statistical models between holistic non-frontal faces and local patches on frontal face.

Besides combining local statistical models with coarse geometrical alignment, another way to integrate geometrical and statistical information is embedding geometrical information into statistical models. Specifically, in this kind of methods, the combination is accomplished by augmenting the feature vector that representing a local region with their spatial locations. And a face is represented as a set of augmented feature vectors. Based on this face representation, Zhao and Gao \cite{cvpr09hausdorff} sampled the augmented feature vectors through key points detection, and used modified Hausdorff distance to measure the similarity between two faces. Wright and Hua \cite{cvpr09JWright} densely sampled the augmented feature vectors on face, and quantified them into histograms via random projection trees. Face recognition is then performed by matching the histograms. Benefiting from the dense sampling and quantization, the matching is spatially elastic to some extent. Thus, the problem of alignment is implicitly alleviated.

The method proposed in this paper is a statistical approach. Base on the foregoing analysis,  it is difficult to get good recognition performance for a pure statistical approach, especially when pose difference is big. Therefore, we extended our method by using some facial landmarks. Similar statistical models are built on the region centered at these landmarks. So, the enhanced recognition method is a hybrid approach that integrates local statistical models with coarse geometrical alignment.
\section{Recognizing Face Across Pose via Canonical Correlation Analysis}
\label{sec_cca}
The problem of recognizing faces with pose difference can be formulated as measuring the similarity between two vectors from different linear subspaces. We propose that it can be solved by Canonical Correlation Analysis. In this section, we describe the CCA in detail and how it leads to pose-invariance or pose-robustness by learning from the coupled face data. In this section, the CCA based recognition approach is also enhanced by integrating holistic and local facial features.

\subsection{Recognizing Face Across Pose via CCA}
Proposed by Hotelling in \cite{hotelling1936cca}, canonical correlation analysis is a classical technique in statistical learning. It has been widely used in pattern recognition, and in recent years it has also been applied to face analysis. Sun and Chen \cite{amc07soft_label,ivc07lpp_cca} modified the CCA model with soft label and local preserving projection respectively. They applied their methods for frontal face recognition. Ma et al. \cite{icml07dcca} improved the CCA model by maximizing the differences between the within class correlations and between class correlations. Their method is also applied for frontal face recognition. Kim et al. \cite{pami07kim_dcca} proposed a similar improvement for the CCA model, but applied their model for face video analysis. Zheng et al. \cite{tnn06expression} used kernel CCA for recognizing facial expression. Reiter et al. \cite{icpr06cca} and Lei et al. \cite{cvpr08cca} used CCA to reconstruct 3D facial shape. Yang et al. \cite{fg08cca} applied CCA to 2D-3D face matching.
\begin{figure}[h]
\begin{center}
   \includegraphics[width=0.9\linewidth]{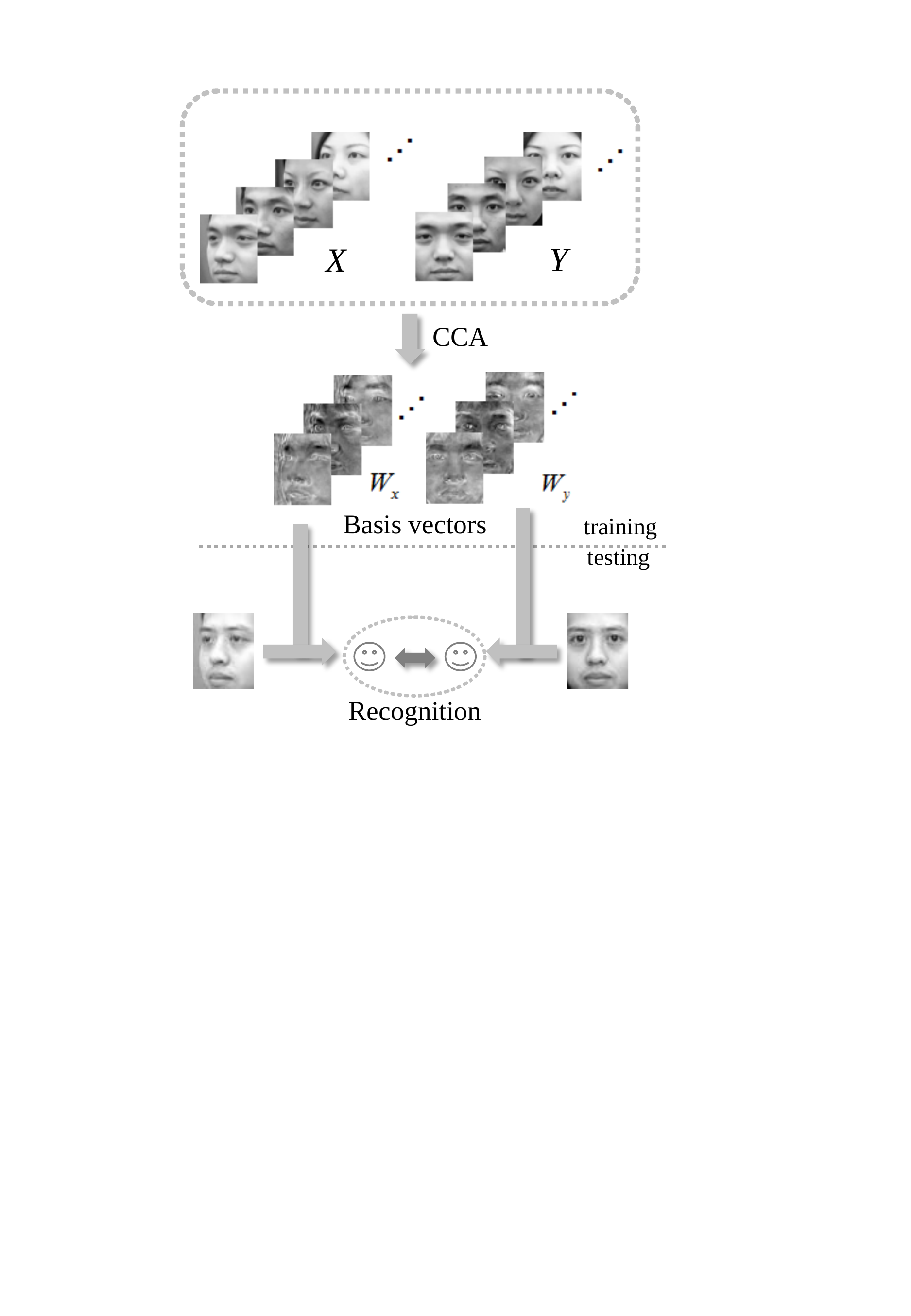}
   \vspace{-3mm}
   \caption{The illustration of our approach. In the training phase two sets of basis vectors are learned via canonical correlation analysis using the coupled face data. In the testing phase faces are projected onto the pose specific basis vectors. Recognition is performed by comparing these projections.}
   \label{fig:flowchart}
   \end{center}
\end{figure}

``Canonical correlation analysis can be seen as the problem of finding basis vectors for two sets of variables such that the correlation between the projections of the variables onto these basis vectors are mutually maximized.''\cite{hardoon2004canonical}. As illustrated in Figure \ref{fig:flowchart}, if the faces from two different poses are coupled by identity and performing CCA on these face pairs, the correlations between the same identities are maximized. Therefore, this intra-subject correlation is robust to pose difference.

Let $(X,Y)$ be the coupled training set of faces from two different poses, where $X = [x_{1},x_{2}\ldots,x_{n}]$, $Y = [y_{1},y_{2}\ldots,y_{n}]$. Each face image is represented by a feature vector, and $n$ is the number of image/vector pairs. Both $X$ and $Y$ are normalized to be of zero mean. Our goal is to find two sets of basis vectors, each for one pose, such that the correlations between the projections of variables onto them are mutually maximized. Denote a pair of basis vectors as $\langle w_{x},w_{y}\rangle$. The correlation $\rho$ between the projections $w^{T}_{x}X$ and $w^{T}_{y}Y$ is
\begin{equation}
\label{eq:cca_obj}
   \rho = \frac{E[w^{T}_{x}XY^{T}w_{y}]}{\sqrt{ E[w^{T}_{x}XX^{T}w_{x}]E[w^{T}_{y}YY^{T}w_{y}]}}.
\end{equation}
Here, $E[f(x,y)]$ is the empirical expectation of function $f(x,y)$.

Considering the means of $X$ and $Y$ are zero, the total covariance
matrix of $(X,Y)$ can be written as:
\begin{equation}
\label{eq:c_matrix}
   C_{total} =  \begin{pmatrix} C_{xx} & C_{xy} \\ C_{yx} & C_{yy} \end{pmatrix} = \frac{1}{n}\begin{pmatrix} X\\Y\end{pmatrix}\begin{pmatrix}
   X\\Y\end{pmatrix}^{T},
\end{equation}
where $C_{xx}$ and $C_{yy}$ are the within-pose covariance matrices
of $X$ and $Y$ respectively and $C_{xy}=C^{T}_{yx}$ is the
within-subject covariance matrix between two different poses.
Thus, the objective function maximizing the correlations can be described as:
\begin{equation}
\label{eq:cca_obj_argmax}
    \langle w_{x},w_{y}\rangle = \arg\max\limits_{\langle w_{x},w_{y}\rangle} \frac{w^{T}_{x}C_{xy}w_{y}}{\sqrt{w^{T}_{x}C_{xx}w_{x}w^{T}_{y}C_{yy}w_{y}}}.
\end{equation}

The solution of $w_{x}$ and $w_{y}$ can be found by solving the
following eigenvalue equations \cite{borga2001canonical}:
\begin{equation}
 \label{eq:cca_eig}
  \begin{aligned}
      & C^{-1}_{xx}C_{xy}C^{-1}_{yy}C_{yx}w_{x} =\rho^{2}w_{x}\\
      & C^{-1}_{yy}C_{yx}C^{-1}_{xx}C_{xy}w_{y} =\rho^{2}w_{y}.
  \end{aligned}
\end{equation}

Only one of the equations needs to be solved, because the solutions are related by
\begin{equation}
\label{eq:eig_relation}
   \begin{aligned}
      & C_{xy}w_{y} =\rho\lambda_{x}C_{xx}w_{x}\\
      & C_{yx}w_{x} =\rho\lambda_{y}C_{yy}w_{y},
   \end{aligned}
\end{equation}
where
\begin{equation}
   \label{eq:lambda}
   \lambda_{x} = \lambda^{-1}_{y} = \sqrt{\frac{w^{T}_{y}C_{yy}w_{y}}{w^{T}_{x}C_{xx}w_{x}}}.
\end{equation}
\begin{figure*}[ht]
\begin{center}
   \includegraphics[width=0.7\linewidth]{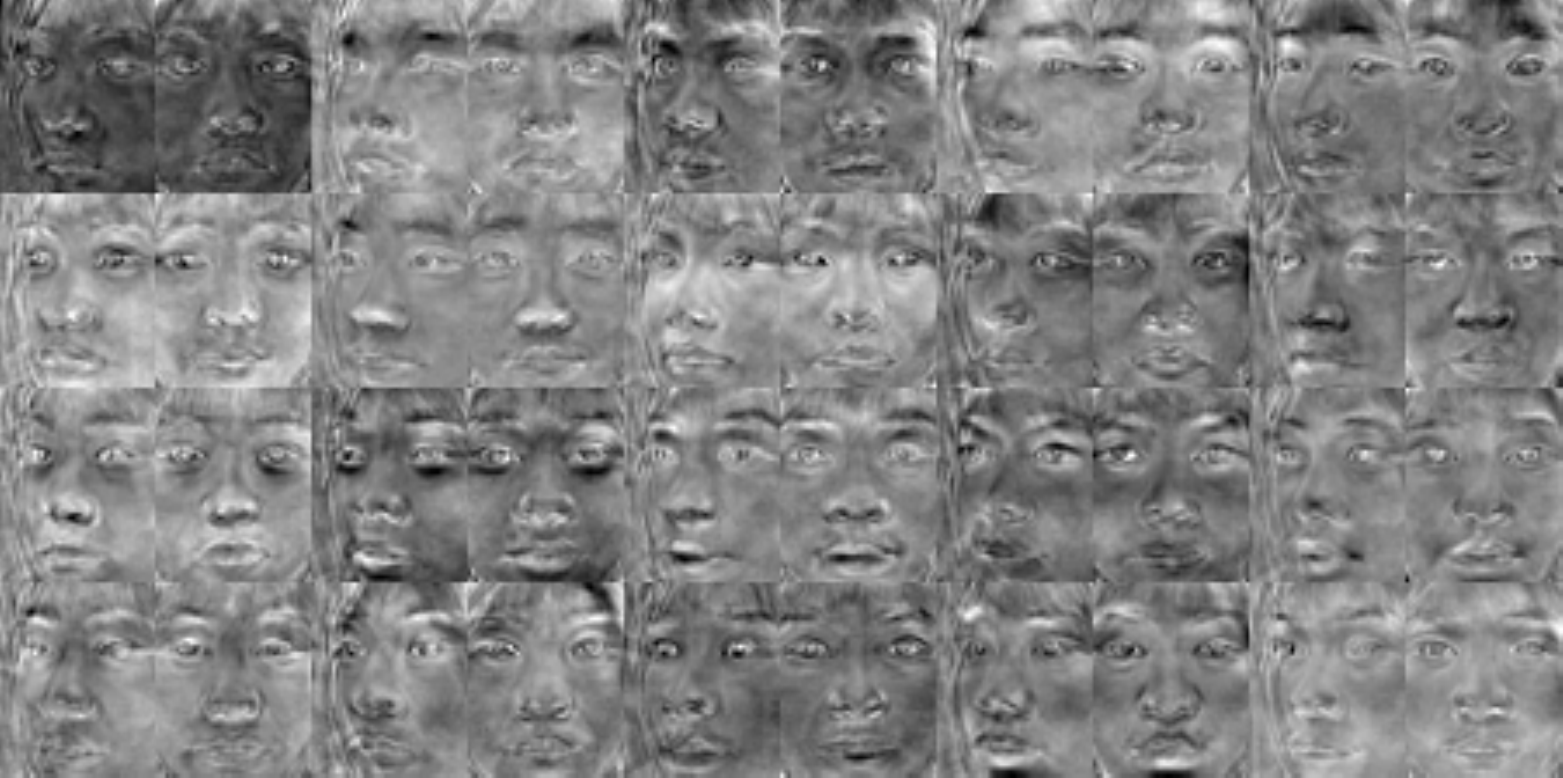}
   \vspace{-3mm}
   \caption{The first 20 \emph{corrfaces} learned via CCA using coupled face data from the CAS-PEAL database with $30^{\circ}$ pose difference.}
   \label{fig:corrfaces}
   \end{center}
\end{figure*}

Eigen-decomposition brings a set of orthogonal eigenvectors. Therefore, we have multiple eigenvector pairs like $\langle w_{x},w_{y}\rangle$. Denote them as $W_{x}=[w_{x1},w_{x2}\ldots,w_{xk}]$ and $W_{y}=[w_{y1},w_{y2}\ldots,w_{yk}]$. Here $k$ is the number of eigenvectors in both sets. Once the optimized $W_{x}$ and $W_{y}$ are obtained, face recognition across poses can be performed by measuring the intra-subject correlations. For a pair of gallery and probe faces $(x_{input},y_{input})$, firstly we project them onto their corresponding basis vectors:
\begin{equation}
   \label{eq:cca_projection}
   \begin{aligned}
   & \hat{x} = W^{T}_{x}(x_{input} - x_{mean}) \\
   & \hat{y} = W^{T}_{y}(y_{input} - y_{mean}).
    \end{aligned}
\end{equation}
The similarity between the gallery and probe face is measured by the correlation between the projections $\hat{x}$ and $\hat{y}$.
\begin{equation}
   \label{eq:corr}
   c = \frac{\hat{x}\cdot\hat{y}}{\|\hat{x}\|\|\hat{y}\|}.
\end{equation}
In the recognition, we simply use the nearest neighbor classifier. A probe face is identified to the gallery face with highest correlation value.

Similar to the Fisher's linear discriminant analysis, singularity problem is also exist in the CCA method, for the covariance matrix $C_{xx}$ and $C_{yy}$ could be not invertible. There are two methods to solve this problem. In the first, covariance matrix can be regularized by adding a small value to the diagonal elements. Denote the small value as $\alpha$, we have:
\begin{equation}
   \label{eq:regular}
   \begin{aligned}
   & C^{*}_{xx} = C_{xx} + \alpha I \\
   & C^{*}_{yy} = C_{yy} + \alpha I.
    \end{aligned}
\end{equation}

By replacing $C_{xx}$ and $C_{yy}$ by the regularized covariance matrix $C^{*}_{xx}$ and $C^{*}_{yy}$ in Equation \ref{eq:cca_eig}, the singularity problem could be avoid. The second method is to perform principal component analysis (PCA) before CCA, which is similar to the fisherfaces method \cite{fisherface97}. Besides solving the singularity problem, PCA also reduces the dimension of the data. Thus, the computational cost in the training phase is also reduced. If the training data is adequate, performing PCA before CCA seems a better way to solve the singularity problem. However, the available multi-pose face data is not adequate enough. For example, in the PIE database \cite{pie_db_pami} there are only 68 subjects. After PCA, the reduced dimension is too low to represent the facial appearance variations. Therefore, in this paper we choose the covariance matrix regularization for solving the singularity problem.

\begin{figure}[h]
\begin{center}
   \includegraphics[width=0.9\linewidth]{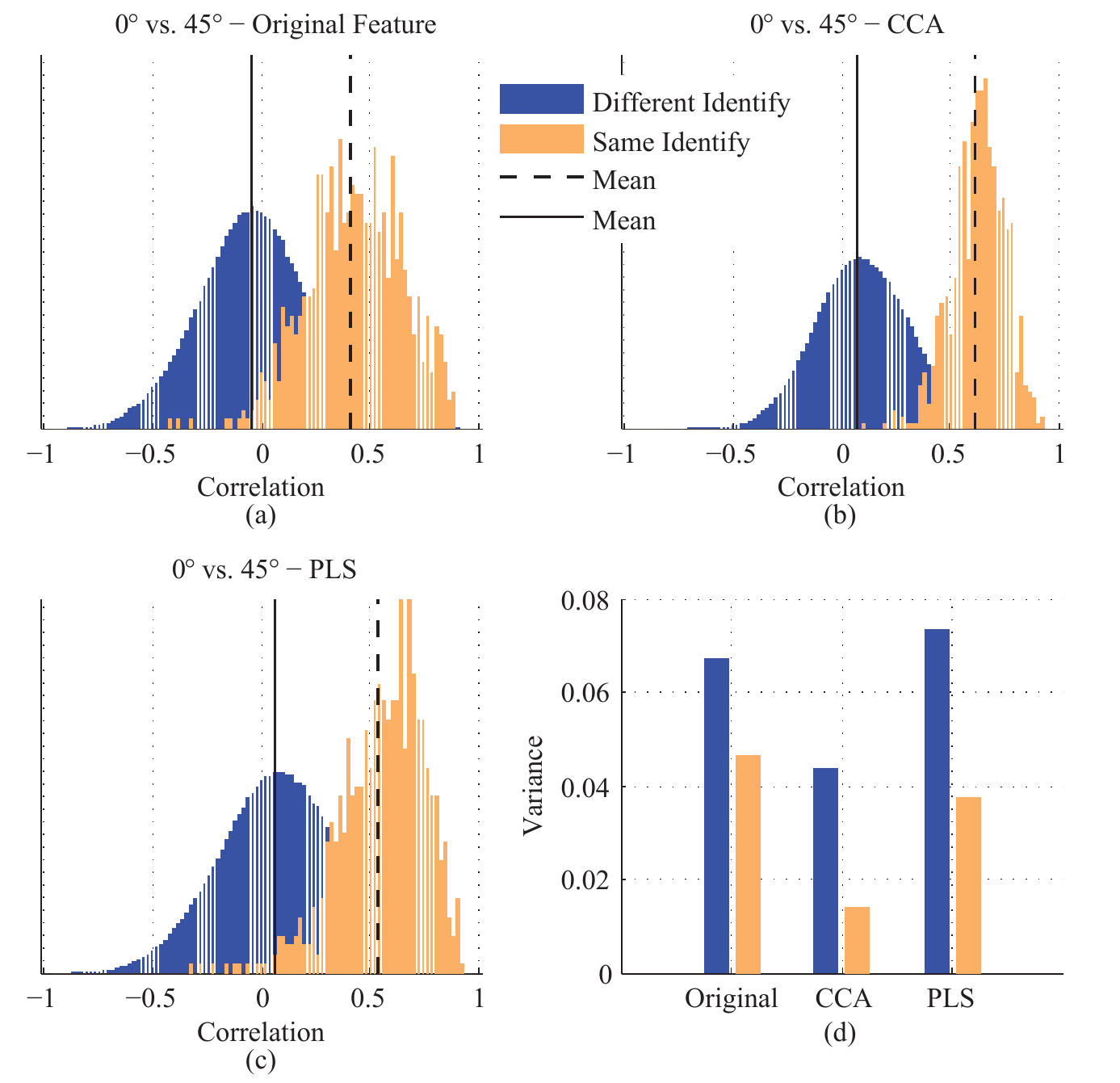}
   \vspace{-3mm}
   \caption{Histograms of correlations before and after CCA/PLS modeling calculated on CAS-PEAL database across $45^{\circ}$ pose difference. In the original feature space (a), correlations between the same and different persons are mixed together. After CCA  and PLS modeling (b,c), the histogram centers (mean value) of the correlations between same the people (the dashed vertical line) move to the right which means that the cross-pose intra-individual correlations are enhanced. The difference between CCA and PLS modeling is that correlations after by CCA modeling are of lower variance (d), which leads to better separation of the histograms.}
      \label{fig:corr_dis}
   \end{center}
\end{figure}

The learning results of CCA are two sets of eigenvectors. Similar to the \emph{eigenfaces} \cite{eigenface91}, we can display these coupled eigenvectors as images. We name these images as \emph{corrfaces}. In Figure \ref{fig:corrfaces}, we presented first 20 corrfaces learned from the CAS-PEAL database with $30^{\circ}$ pose difference. Different from the eigenfaces, the corrfaces are coupled pairs. We can see that in each pair of corrfaces, the ghost faces belong to the same ``ghost''. The eigenvectors $W_{x}$ and $W_{y}$ can be viewed as two coordinate systems that reflect the same identity information in different pose specific subspaces. Projecting the faces of different pose onto $W_{x}$ and $W_{y}$ respectively, the influence of identity is emphasized while the influence of pose difference is reduced. In Figure \ref{fig:corr_dis}, we demonstrate the learning results on CAS-PEAL database by comparing the histograms of the correlation values. In Figure \ref{fig:corr_dis} (a), we present the histogram of the correlations between original feature vectors. Projecting the original feature vectors onto the corrfaces, the histogram of their correlations is shown in Figure \ref{fig:corr_dis} (b). We can see that the in Figure \ref{fig:corr_dis} (b) the correlation distributions of the same and different identity are better separated. The influence of pose misalignment in the feature vectors is considerably reduced.
\subsection{Comparisons with Partial Least Squares}
\label{subsec:pls_cca}
Partial Least Squares (PLS) is similar to CCA in modeling correlations between two different sets. The relations between them have been well studied. In this work, we compare them by following the formulation in \cite{borga2001canonical}. The eigen-problems in Equation \ref{eq:cca_eig} can be formulated as a single eigenvalue equation:
\begin{equation}
   \label{eq:cca_pls_eig}
   B^{-1}A\hat{w} = \rho\hat{w},
\end{equation}
where
\begin{equation}
   \label{eq:cca_pls_ab_matrix}
   A = \begin{pmatrix} 0 & C_{xy} \\ C_{yx} & 0 \end{pmatrix}, B =\begin{pmatrix} C_{xx} & 0 \\ 0 & C_{yy} \end{pmatrix}, \hat{w} = \begin{pmatrix} \mu_{x}\hat{w}_{x} \\ \mu_{y}\hat{w}_{y} \end{pmatrix}.
\end{equation}
Equation \ref{eq:cca_pls_eig} provides a unified framework of solving PCA, PLS and CCA by different matrices A and B, which is described in Table \ref{tab:pls_cca}.
\begin{table}
\caption{The matrices A and B for PCA, PLS and CCA.}
\label{tab:pls_cca}
\centering \small
\begin{tabular}{|c|c|c|}
\hline
Model & A & B \\ 
\hline
PCA & $C_{xx}$ & I \\ 
\hline
PLS &  $\begin{pmatrix} 0 & C_{xy} \\ C_{yx} & 0 \end{pmatrix}$ & $\begin{pmatrix} I & 0 \\ 0 & I \end{pmatrix}$ \\ 
\hline
CCA &  $\begin{pmatrix} 0 & C_{xy} \\ C_{yx} & 0 \end{pmatrix}$ & $\begin{pmatrix} C_{xx} & 0 \\ 0 & C_{yy} \end{pmatrix}$ \\ 
\hline
\end{tabular}
\end{table}

As can be seen, the difference between PLS and CCA is the choice of matrix B, which corresponds to the normalization denominator in Equation \ref{eq:cca_obj_argmax}. In another word, the normalization on correlations is the key difference between PLS and CCA modeling. In Figure \ref{fig:corr_dis}, we illustrate the results of PLS and CCA modeling obtained from CAS-PEAL database with $45^{\circ}$ pose difference. After PLS and CCA modeling, the distribution centers (dashed vertical lines) of intra-individual correlations both move to the right, which means their values are maximized. This phenomenon reflects the influence of matrix A and the numerator in Equation \ref{eq:cca_obj_argmax}, which is the same for PLS and CCA. As shown in Figure \ref{fig:corr_dis} (d), the results of CCA modeling are  lower in variance, which demonstrates the influence of matrix B and the normalization denominator in Equation \ref{eq:cca_obj_argmax}. Because of the lower variance, CCA separates the histograms better than PLS does. It shows that the variance of intra-individual correlations is high in cross-pose face recognition, and the normalization in CCA reduces the variance and gives an extra contribution to the improvement of performance.

\subsection{Enhancement with holistic+local feature representation}

\begin{figure}[h]
\begin{center}
   \includegraphics[width=0.7\linewidth]{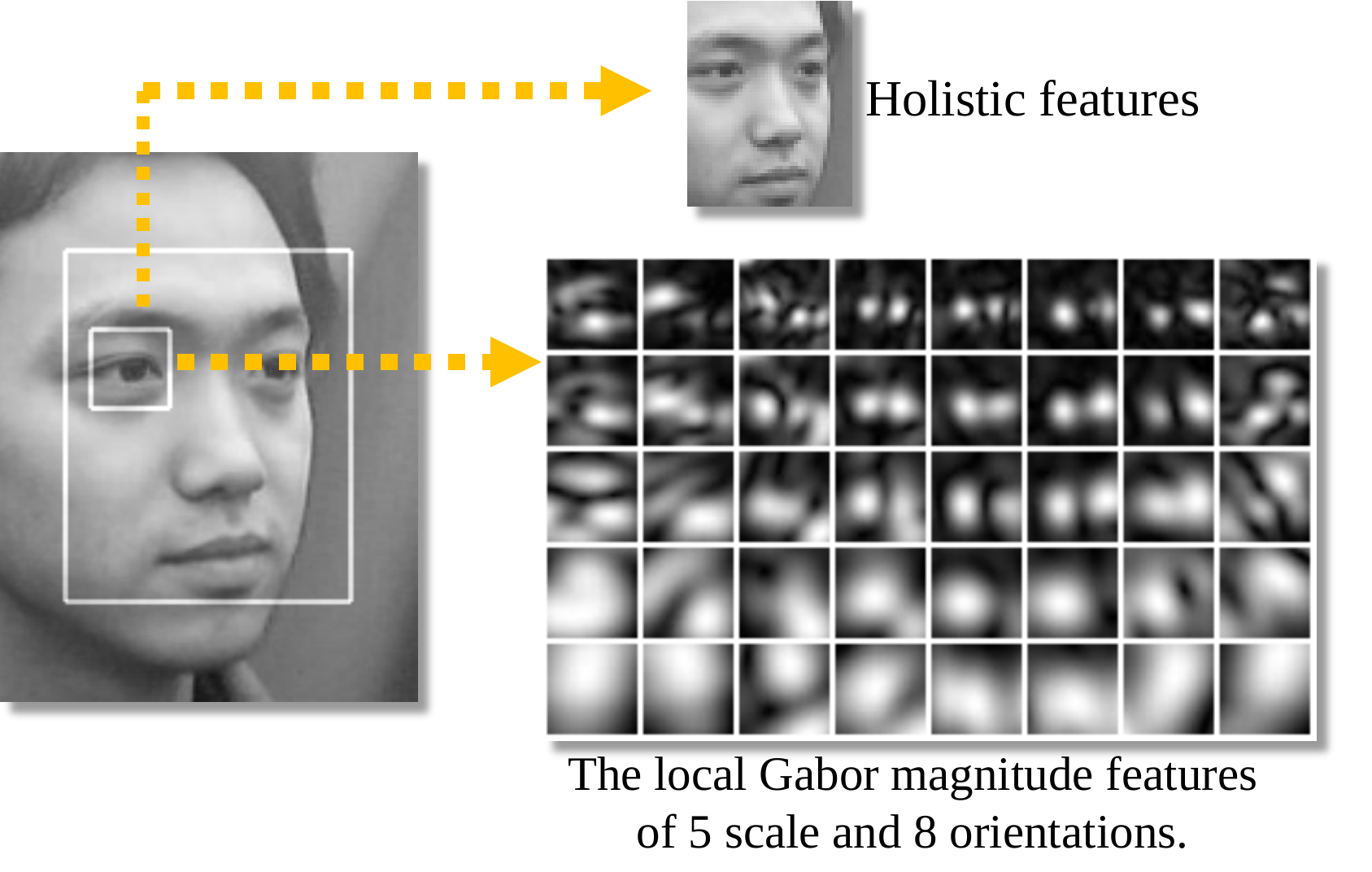}
   \caption{The holistic and local facial feature representation. The holistic features are sampled from holistic face region. The local Gabor magnitude features of 5 scale and 8 orientations are extracted from local region centered at facial landmarks, such as the eye centers.}
\label{fig:feat_rep}
   \end{center}
\end{figure}

The pixel intensity based feature vector is frequently used in the face recognition literature. However, its representation power is limited. The 2D Gabor features were proved successful and widely used in face recognition \cite{pami97EBGM,TIP02gabor_liu}. But if the Gabor features are sampled from the whole face, such as the approach of Liu and Wechsler \cite{TIP02gabor_liu}, the pose misalignment in feature vectors would counteract the benefit of the Gabor features. Since the Gabor wavelet is a 2D texture descriptor, while pose variations are tightly related to the 3D shape of human face. Convoluted by the Gabor kernel, the geometrical distortion in the feature vectors becomes even worse than that in the pixel intensity based feature vectors. This problem could be avoided by using coarse geometrical alignment. Prince et al. \cite{pami08tiedfactor} extract local Gabor features on some local regions centered at some corresponding facial landmarks, such as the center of eye, the corner of mouth etc. This feature representation approach considerably improves the recognition performance. To explore the potentiality of the proposed method, we adopte similar feature representation method. One difference is that besides the local Gabor features, the holistic features are also used in our method. The local Gabor features mainly reflect the local information of face. Su et al. \cite{SuYu_tip09} proved that the holistic features are also important for representing faces. In their work, classifiers are built on the holistic features and local Gabor features respectively. By combining the holistic and local classifiers, the performance of frontal face recognition is dramatically improved. In this paper we also use the holistic+local strategy for feature enhancement. As shown in Figure \ref{fig:feat_rep}, the holistic intensity features are sampled on the whole face region, while the local Gabor magnitude features of 5 scale and 8 orientations are extracted on the patches centered at some facial landmarks, such as the eye centers, the corner of mouth etc. It should be pointed out that the enhancement in feature representation benefits from both the coarse alignment and the Gabor features. Enhanced by this feature representation method our approach becomes a hybrid approach that integrates coarse geometric alignment and local statistical models.

Based on the approach described above, we have several independent models for representing the faces. Besides the holistic model, we have several local models each local model of which corresponds to a facial landmark. There are two different ways to utilize these models, i.e., concatenating these models into a holistic feature vector and using them independently. In this paper we choose to use these models independently for reducing the computational cost. Consequently, a CCA based classifiers is built independently on each holistic model and local model. Denote the correlation given by the \emph{i-th} classifier as $c_{i}$, the final decision is derived from the mean correlation calculated by the following equation:
\begin{equation}
\label{eq:s_total}
   S_{total} = \frac{1}{k}\sum\limits_{i=1}^{k}c_{i}.
\end{equation}

\section{Experiments}
\label{sec_exp}
In order to validate the proposed approach of pose-robust face recognition, we performed experiments on three databases, i.e., the CMU PIE \cite{pie_db_pami}, the Multi-PIE \cite{icv09multi_pie} and the CAS-PEAL database \cite{cas_peal_db} respectively. In the PIE database there are 68 people whose images are captured in 13 poses with yaw and pitch angle differences. The yaw angle difference between neighbor poses is about 22.5$^\circ$. The Multi-PIE database is an extended version of the PIE. It contains more subjects. In this database face images are captured under 15 viewpoints in four recording sessions. In our experiments, we use face data from session 1, which contains 249 subjects. The interval of neighbor yaw angle is about 15$^\circ$. In the CAS-PEAL database, the subject is asked to look upward, downward and horizontally for capturing faces under different pitch angles. In our experiments we use the face images under horizontal pitch angle, which consists of face images of 938 subjects under 7 different yaw angles. The yaw angle difference between neighbor poses is about 15$^\circ$.
\begin{figure}[h]
\begin{center}
   \includegraphics[width=0.45\linewidth]{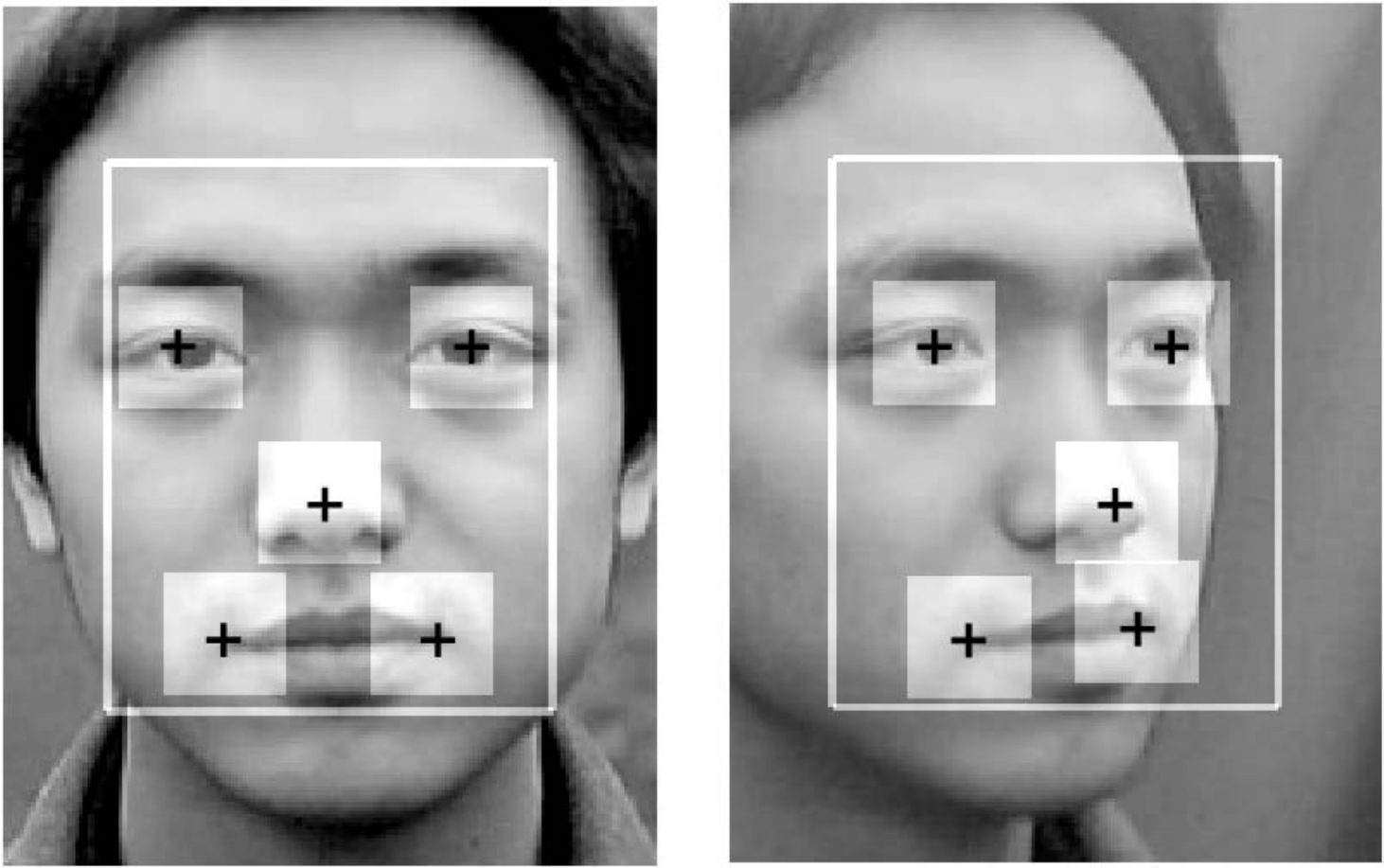}
   \vspace{-3mm}
   \caption{The facial landmarks used in the experiments.}
   \label{fig:landmarks}
   \end{center}
\end{figure}

Five landmarks are manually labeled on each face in our experiments, i.e. the eye centers, the tip of nose and the corners of mouth respectively. The face image is normalized using the eye centers and mouth corners. The scale of face is normalized according to the distance between the center of mouth and the midpoint of two eye centers. All the face images are normalized to $204\times256$. The example of facial landmarks and normalized faces are illustrated in Figure \ref{fig:landmarks}.
\begin{figure*}[ht]
\begin{center}
   \includegraphics[width=0.86\linewidth]{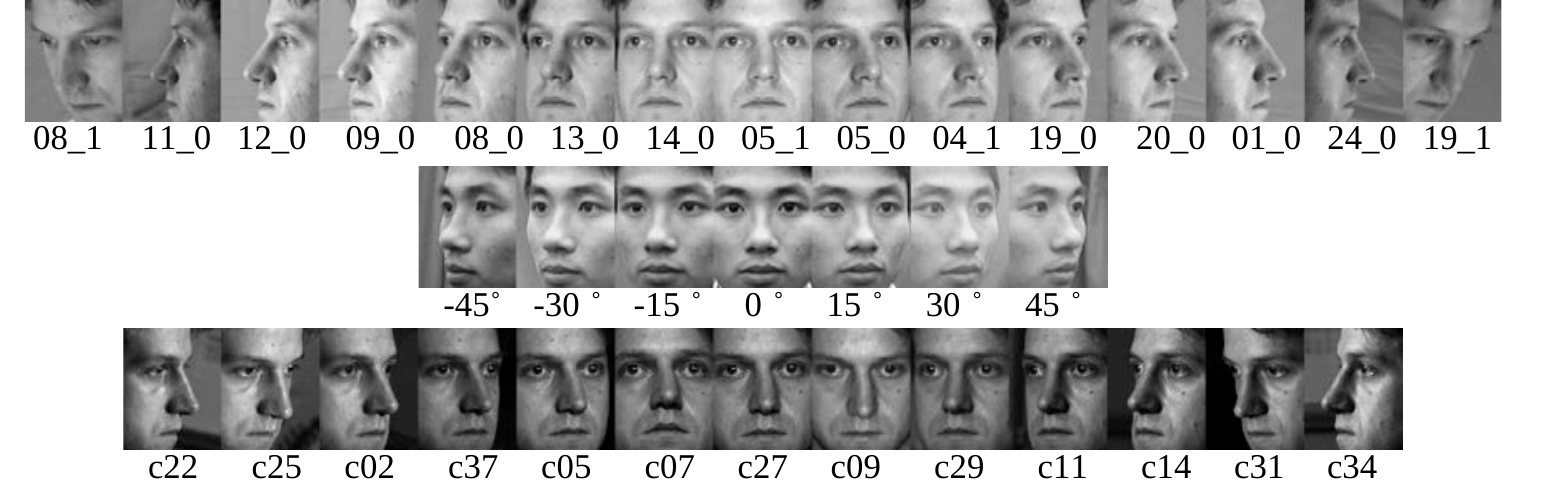}
   \vspace{-3mm}
   \caption{Examples of normalized holistic face region of Multi-PIE (top row), CAS-PEAL (middle row) and PIE (bottom row) database.}
   \label{fig:db_example}
   \end{center}   
\end{figure*}
\begin{figure*}[ht]
\begin{center}
   \includegraphics[width=0.7\linewidth]{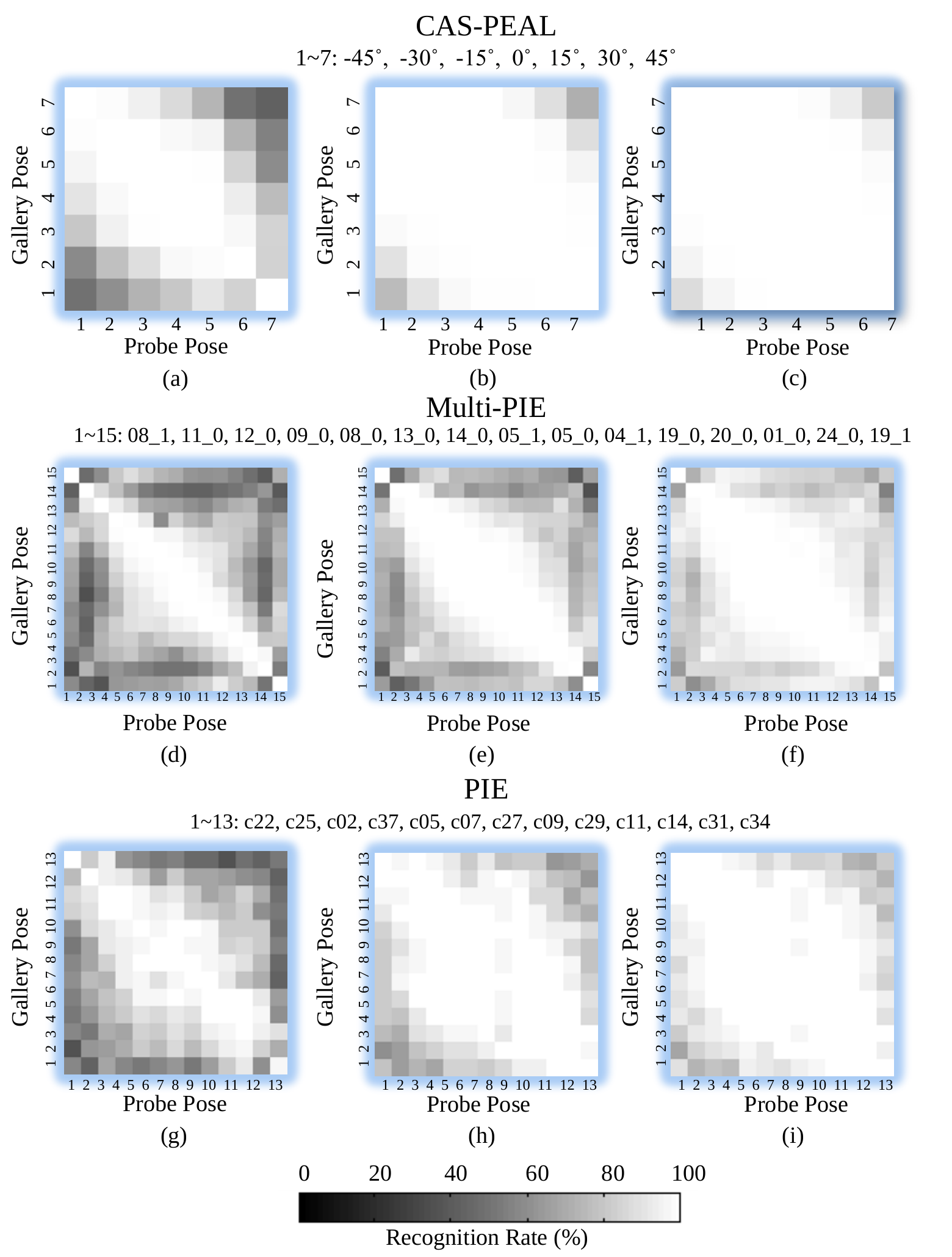}
   \vspace{-3mm}
   \caption{Experimental results on CAS-PEAL (top row), Multi-PIE (middle row) and PIE (bottom row) database with holistic (a,d,g), local (b,e,h) and holistic+local (c,f,i) feature representation.}
\label{fig:results_matrix}
   \end{center}
\end{figure*}

We performed three series of experiments. In experiment 1, the proposed method is evaluated on CAS-PEAL and Multi-PIE database. Although CAS-PEAL and Multi-PIE are better benchmark for cross-pose face recognition, previous approaches are not tested on them. Therefore, we compare the proposed method with related approaches on PIE in experiment 2. In experiments 1 and 2, the pose angle of gallery and probe faces is assumed to be known. To evaluate the robustness to inaccurate pose estimation, we perform experiments with unknown probe pose in experiment 3.

\subsection{Experiment 1: Comparisons on CAS-PEAL and Multi-PIE}

In this experiment, we test the proposed method with three types of visual features, i.e. the holistic intensity feature, the local Gabor feature and the combination of them. In these experiment the normalized holistic face region (the rectangle in Figure \ref{fig:landmarks}) is re-sized to $45\times56$. Thus, the length of the holistic feature vector is 2520. Some examples of normalized holistic face region are shown in Figure \ref{fig:db_example}. Although holistic intensity feature is frequently used in the face recognition research, its representation power is limited. To evaluate the potentiality of the proposed method, we align the local face region with some facial landmarks and used Gabor filters for features extraction. As shown in Figure 8, we use five facial landmarks, i.e. two centers of eyes, two corners of mouth and the tip of nose respectively. These landmarks are the most salient points on face. Unlike these points on the silhouette or close to the hair, they are robust to occlusion and easy to detect. The Gabor magnitude features of 5 scales and 8 orientations are sampled from a  $31\times31$ window centered at each landmark. To reduce the dimension, the window is down sampled to $7\times7$. Thus, the total length of Gabor feature vector is 1960.

For experiments on Multi-PIE database, we used 100 subjects for training and 149 subjects for testing. In the experiments on CAS-PEAL database, 200 subjects are used for training and the remaining 738 subjects are used for testing. We find that the bigger number of basis vectors the better performance CCA achieves. Therefore, subtracting the one dimension of the data centering (zero mean), we set the number of basis vectors to 99 and 199 for Multi-PIE and CAS-PEAL respectively\footnote{We simply use maximum dimensions for preserving energy as more as possible. For the training data is centered by subtracting the mean face, the maximum dimension is $n$-1, where $n$ is the number of training samples.}. Moreover, the regularization parameter $\alpha$ in Equation \ref{eq:regular} is set to $10^{-6}$ in all experiments.

For the holistic faces, we build a CCA based classifier. And for local features, a CCA based classifier is built independently on each pair of local regions centered at the corresponding landmarks. Consequently, we have 1 holistic classifier and 5 local classifiers in total. The final classification decision is obtained by integrating these classifiers.  

We plot the matrix of recognition results (all poses against all poses) with holistic and local features in Figure \ref{fig:results_matrix}. The experimental results on CAS-PEAL with holistic feature, local feature and holistic+local features are given in Figure \ref{fig:results_matrix} (a), (b) and (c) respectively. Corresponding average recognition rates are  85.73\%, 97.49\% and 98.72\%. In Figure 10 (d), (e) and (f) we illustrate the results on Multi-PIE. The average recognition rates with holistic feature, local feature and the fusion of them are 73.31\%, 82.96\% and 91.31\% respectively.  

To illustrate the relationship between performance and pose difference, we show the performance comparison under frontal gallery pose in Figure \ref{fig:results_curve}. As can be seen, combining 5 local classifiers could get much higher performance than single holistic classifier. The CAS-PEAL database contains the largest number of subjects in multi-pose databases. It is impressive that, when the galley pose is frontal, over 99\% recognition rates are achieved in all probe poses on the CAS-PEAL database. As we analyzed in Section \ref{sec_intro}, the pose difference makes the holistic feature vector noisy and misaligned. Compared with holistic feature aligned local features are more robust to pose variations. Thus, representing face locally could improve the performance. However it does not mean that the holistic feature is useless. When pose difference is big the alignment becomes inaccurate. By contrast, in this scenario holistic features are more robust than the local features. Our experimental results show that combining holistic features could improve the performance in cases with large pose differences. Since the pose differences in the experiments on Multi-PIE are larger, the performance enhancement is clearer than that on PIE and CAS-PEAL. From Figure \ref{fig:results_curve}, we can see that the larger the pose difference is, the more performance enhancement we can get. Integrating holistic features can considerably improve the performance for cross-pose face recognition.

In Figure \ref{fig:results_curve}, we also compared the proposed cross-pose face recognition method with partial least squares (PLS) \cite{prl2011annan}, ridge regression \cite{tip2012annan} and the tied factor analysis (TFA) \cite{pami08tiedfactor}\footnote{The implementation of TFA is based on the codes provided in \url{http://web4.cs.ucl.ac.uk/staff/s.prince/TiedFactorAnalysis.zip}.}. In the experiments, the number of tied factors is set to 32. We can see that CCA outperforms PLS and ridge regression using the same visual features. As described in Section \ref{subsec:pls_cca} CCA is better than PLS in controlling the variance of intra-individual correlations. As a result, it archives better performance than PLS does. Both using holistic features, CCA outperforms TFA on CAS-PEAL. Among the 14 probe poses in Multi-PIE, CCA and TFA achieve the same accuracy in 1 pose, and TFA is better in 8 poses, while CCA outperforms TFA in 5 poses. 

\subsection{Experiment 2: Comparisons on PIE }

The Multi-PIE and CAS-PEAL database are recently released. Compared with PIE they contain more subjects. Especially in our experiments on CAS-PEAL database 738 subjects are used for testing, which is a much bigger number than in previous experiments. The large-scale of data sets makes experimental results more convincing. However, the experiments in previous works of cross-pose face recognition are not performed on Multi-PIE and CAS-PEAL. For comparison to previous works, we also conduct experiments on the PIE database.

In the PIE database there are 68 people whose images are captured in 13 poses with yaw and pitch angle differences. The yaw angle difference of neighbor pose is about 22.5$^\circ$. In the experiments one half of the data are used for training while the remaining part is used for testing. In this experiment the gallery and probe pose are known. The feature representation and parameter settings are the same with that in experiments on Multi-PIE and CAS-PEAL.

\begin{figure}[ht]
\begin{center}
   \includegraphics[width=0.99\linewidth]{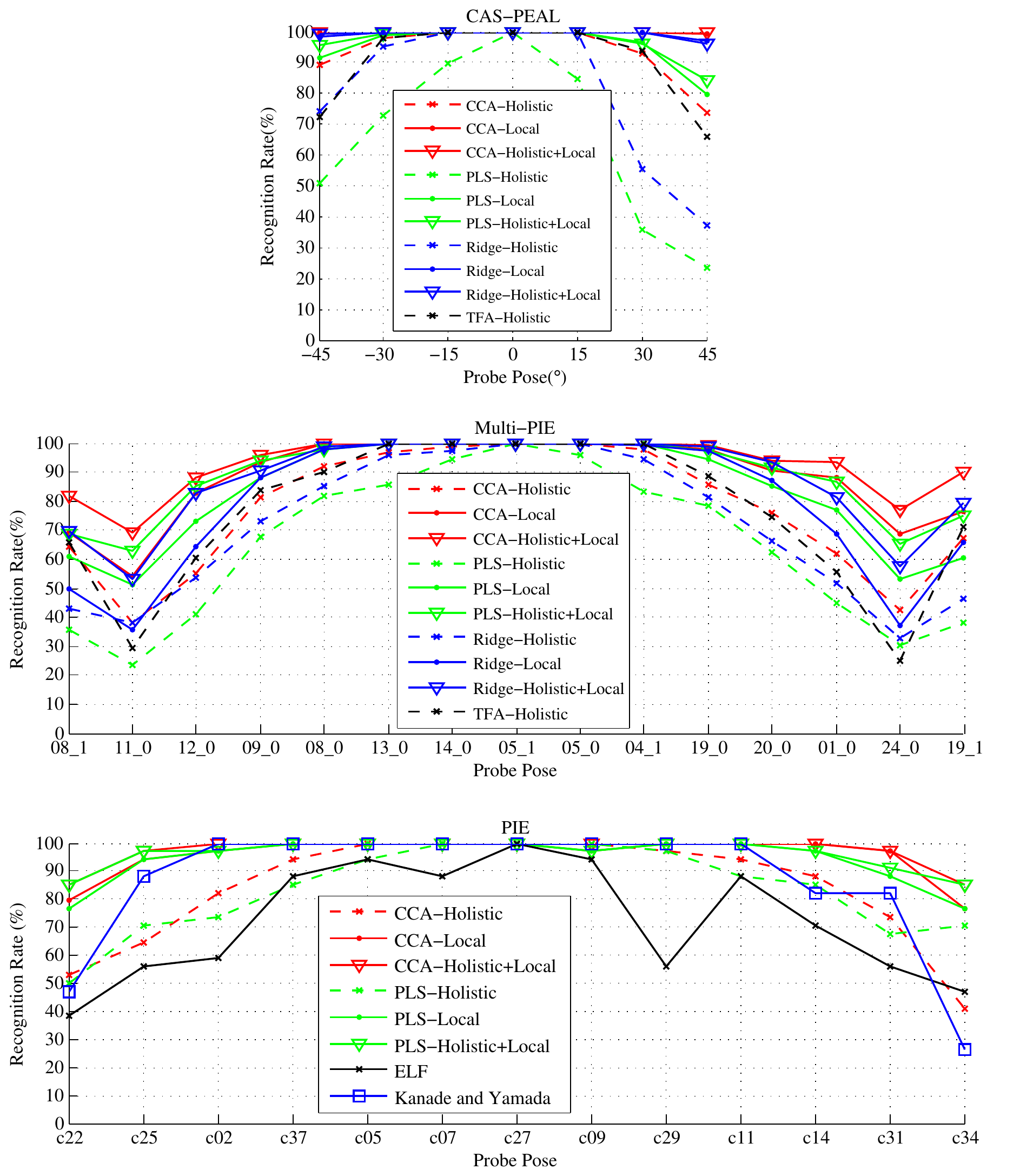}
   \vspace{-3mm}
   \caption{Experimental results of cross-pose face recognition using frontal gallery faces and non-frontal probe faces on CAS-PEAL,PIE and Multi-PIE data sets.}
   \label{fig:results_curve}
   \end{center}
\end{figure}

The experimental results are given in the bottom row of Figure \ref{fig:results_matrix}. The average recognition rates with holistic feature, local feature and the fusion of them are 77.46\%, 91.98\% and 95.28\%. In Figure \ref{fig:results_curve}\footnote{In Figure \ref{fig:results_curve}, we directly cite the experimental results reported in \cite{PAMI04ELF,KanadeYamada03}.} we show the experimental results using frontal gallery face with comparisons of Eigen-light field proposed by Gross et al. \cite{PAMI04ELF} and the multi-subregion matching method proposed by Kanade and Yamada \cite{KanadeYamada03}. All the three methods shown in Figure \ref{fig:results_curve} use 34 subjects for testing. Thus the comparison is relatively fair. If we only use the holistic intensity features, our method outperforms the Eigen-light field in most viewpoints, which also represents face holistically. The performance of our method with holistic features is lower than the Kanade and Yamada's approach, which is a local patch based approach. But when local Gabor features are used, the performance of our method is better.
\begin{table}
\caption{Performance comparison on CMU PIE database.}
\label{tab:pie_compare}
\centering \scriptsize
\begin{tabular}{ccc}
Work &  Pose diff. ($^\circ$) & Rec. rate ($\%$) \\ \hline
Blanz and Vetter \cite{pami03_3dmm} &  front / side / profile & 99.8 / 97.8 / 79.5 \\ 
Prince et al. \cite{pami08tiedfactor} &  22.5 / 90 & 100 / 91 \\ 
Chai et al. \cite{TIP07LLR} &  22.5 / 45 & 98.5 / 89.7 \\ 
Kanade and Yamada \cite{KanadeYamada03} & 22.5 / 45 / 90  & 100 / 100 / 47 \\ 
Gross et al.  \cite{PAMI04ELF} (ELF Complex)&  22.5 / 45 / 90 & 93 / 88 / 39 \\ 
Castillo and Jacobs \cite{pami09stereo_matching} (4ptSMD)& 22.5 / 45 / 90 & 100 / 97 / 62 \\
Sharma and Jacobs \cite{cvpr11pls} & 22.5 / 45 / 90 & 100 / 88 / 79 \\
Asthana et al. \cite{iccv11full_auto} & 22.5 / 45  & 100 / 98.5 \\
Mostafa et al. \cite{eccv12at_distance} & 22.5 / 45  & 100 / 95.6 \\
\hline
Our method & 22.5 / 45 / 90 & 100 / 100 / 85.29 \\ \hline
\end{tabular}
\end{table}

To our knowledge, the 3D morphable model (3DMM) proposed by Blanz and Vetter \cite{pami03_3dmm} and the tied factor analysis (TFA) proposed by Prince et al. \cite{pami08tiedfactor} are the state of the art in cross-pose face recognition. For comparing the results reported in \cite{pami08tiedfactor}, empirical comparisons in three probe views (c05,c11,c22) on PIE database are given in Table \ref{tab:pie_compare}\footnote{The experimental results in Table \ref{tab:pie_compare} except our method are cited from corresponding references.} (The gallery pose is frontal). Besides the 3DMM and TFA, other notable results on PIE are also illustrated. From the table, we can see that TFA achieves the best results, the performance of our method using holistic+local features is between 3DMM and TFA. It should be pointed out that the comparisons are empirical for experiment settings and feature representation methods are different. In the experiments of TFA and our method, only half of the subjects are used in testing. In the experiments of 3DMM all the subjects are included in the testing set. The number of facial landmarks used in the experiments influences the power of feature representation. In our experiments we use only 5 landmarks, while 6-8 and 14 manually labeled landmarks are used in the experiments of 3DMM and TFA respectively. It could be expected that the performance of our method could be further enhanced by using more facial landmarks. Empirically speaking, the proposed method is close to the state of the art.

\subsection{Experiment 3: When probe pose is unknown}
The probe pose is assumed known in experiment 1 and 2. To evaluate the robustness to inaccurate head pose estimation of the proposed method, we conduct experiments on the CAS-PEAL database with unknown probe pose. Similar to the experiment 1 and 2, face images of 200 subjects are used for training, while the remaining 738 subjects are used for testing. The gallery pose is frontal. The probe pose is non-frontal and unknown. Therefore, there are six possible poses in total. The experimental results with holistic, local and ``holistic+local'' feature representation are plotted into $6\times6$ ``confusion matrices'' in Table \ref{tab:unknown_pose_holistic}, \ref{tab:unknown_pose_local} and \ref{tab:unknown_pose_fusion} respectively. In these tables ``EP'' denotes the estimated probe poses and  ``RP'' denotes the real probe poses. In the experiments the facial landmarks are also manually labeled.

\begin{figure}[ht]
\begin{center}
   \includegraphics[width=0.8\linewidth]{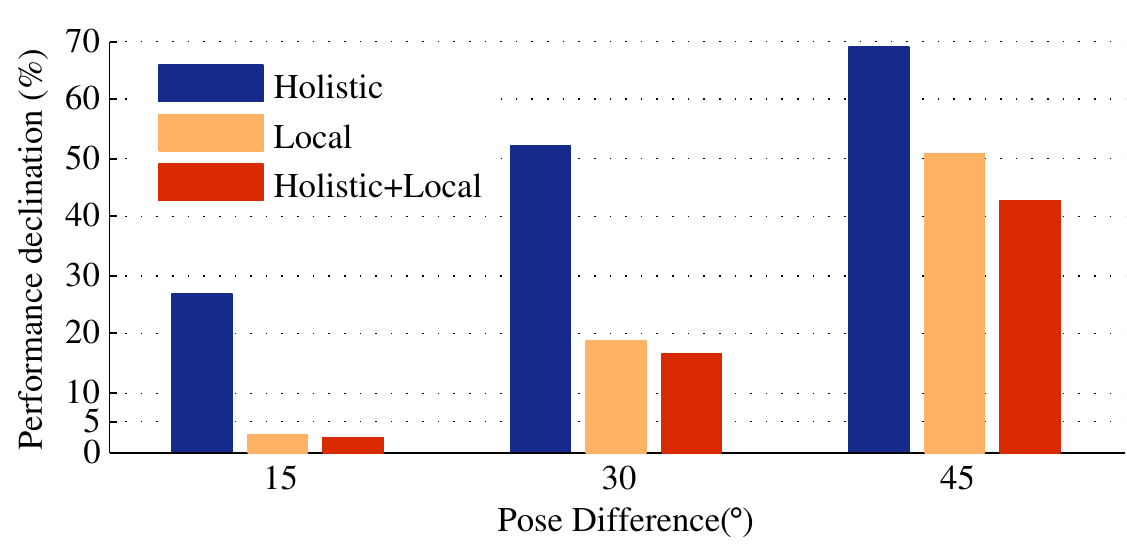}
   \vspace{-3mm}
   \caption{The average performance declination.}
   \label{fig:unknown_pose}
   \end{center}
\end{figure}

As can be seen, the local feature representation are more robust to inaccurate pose estimation. Besides the alignment and feature representation, the recognition performance is also affected by the difference between gallery pose and real probe pose. The smaller the difference the higher the performance is. In Figure \ref{fig:unknown_pose} we illustrate the average performance declination when the differences between the estimated pose and the real pose are $\pm15^{\circ}$,$\pm30^{\circ}$ and $\pm45^{\circ}$. We can see that it is still difficult for recognizing faces across pose when the error of pose estimation is big. When the error of pose estimation is about $15^{\circ}$ the average decrease in performance is 2.50\% based on ``holistic+local'' feature representation.  We can conclude that the proposed method is robust under this condition. From recent survey \cite{pami09pose_estimation}, achieving error less than $15^{\circ}$ is not difficult for pose estimation technique.
\begin{table}[ht]
\caption{The confusion matrix of recognition rate based on holistic features.}
\label{tab:unknown_pose_holistic}
\centering \scriptsize
\centering
\begin{tabular}{|c|c|c|c|c|c|c|}
  \hline
     \backslashbox{EP\kern-2em}{RP} & $-45^{\circ}$ & $-30^{\circ}$ & $-15^{\circ}$ & $15^{\circ}$ & $30^{\circ}$ & $45^{\circ}$ \\ \hline
  $-45^{\circ}$ & 88.75\% & 52.57\% & 36.72\% & 22.76\% & 12.46\% & 7.31\% \\ \hline
  $-30^{\circ}$ & 66.80\% & 94.17\% & 85.36\% & 45.79\% & 17.88\% & 9.07\% \\ \hline
  $-15^{\circ}$ & 31.30\% & 70.32\% & 94.44\% & 51.62\% & 21.40\% & 7.31\% \\ \hline
  $15^{\circ}$ & 20.86\% & 52.84\% & 87.12\% & 100\% & 57.18\% & 22.08\% \\ \hline
  $30^{\circ}$ & 16.53\% & 36.58\% & 67.88\% & 87.66\% & 90.51\% & 40.92\% \\ \hline
  $45^{\circ}$ & 11.11\% & 18.56\% & 26.01\% & 43.22\% & 54.60\% & 75.88\% \\
  \hline
\end{tabular}
\end{table}
\begin{table}[ht]
\caption{The confusion matrix of recognition rate based on local features.}
\label{tab:unknown_pose_local}
\centering \scriptsize
\centering
\begin{tabular}{|c|c|c|c|c|c|c|}
  \hline
   \backslashbox{EP\kern-2em}{RP} & $-45^{\circ}$ & $-30^{\circ}$ & $-15^{\circ}$ & $15^{\circ}$ & $30^{\circ}$ & $45^{\circ}$ \\ \hline
  $-45^{\circ}$ & 100\% & 99.18\% & 98.50\% & 49.45\% & 18.97\% & 13.55\% \\ \hline
  $-30^{\circ}$ & 92.41\% & 99.86\% & 100\% & 69.91\% & 49.05\% & 31.57\% \\ \hline
  $-15^{\circ}$ & 85.77\% & 99.05\% & 100\% & 95.12\% & 75.33\% & 45.25\% \\ \hline
  $15^{\circ}$ & 38.61\% & 64.36\% & 91.86\% & 99.86\% & 98.23\% & 77.64\% \\ \hline
  $30^{\circ}$ & 29.67\% & 46.47\% & 79.40\% & 99.86\% & 100\% & 94.85\% \\ \hline
  $45^{\circ}$ & 18.83\% & 33.33\% & 65.31\% & 99.05\% & 99.05\% & 99.45\% \\
  \hline
\end{tabular}
\end{table}
\begin{table}[ht]
\caption{The confusion matrix of recognition rate based on ``holistic+local'' feature representation.}
\label{tab:unknown_pose_fusion}
\centering \scriptsize
\centering
\begin{tabular}{|c|c|c|c|c|c|c|}
  \hline
   \backslashbox{EP\kern-2em}{RP} & $-45^{\circ}$ & $-30^{\circ}$ & $-15^{\circ}$ & $15^{\circ}$ & $30^{\circ}$ & $45^{\circ}$ \\ \hline
  $-45^{\circ}$ & 100\% & 99.45\% & 98.50\% & 60.97\% & 25.74\% & 16.12\% \\ \hline
  $-30^{\circ}$ & 96.34\% & 99.86\% & 100\% & 78.86\% & 51.62\% & 32.65\% \\ \hline
  $-15^{\circ}$ & 84.41\% & 99.18\% & 100\% & 94.03\% & 71.40\% & 43.22\% \\ \hline
  $15^{\circ}$ & 38.88\% & 67.20\% & 95.52\% & 100\% & 97.83\% & 74.66\% \\ \hline
  $30^{\circ}$ & 40.78\% & 67.61\% & 91.73\% & 100\% & 100\% & 92.41\% \\ \hline
  $45^{\circ}$ & 34.95\% & 50.54\% & 79.53\% & 99.59\% & 99.59\% & 99.72\% \\
  \hline
\end{tabular}
\end{table}
\section{Conclusions and future works}
\label{sec_conclusions}

In this paper we proposed a novel approach for recognizing faces across different poses. We showed that it is the misalignment in feature vectors that makes the cross pose face recognition very difficult. However, by learning via canonical correlation analysis from the face pairs coupled by the same identities, the intra-subject correlations could be maximized across different poses. Thus, the problem of matching misaligned vectors is statistically avoided. We conducted experiments on the largest and latest multi-pose data sets. The experimental results show that the proposed approach is very effective. Our experiments also show that integrating local and holistic features can further improve the recognition performance, especially when the pose difference is large and thus the information of local appearance is unreliable because of the occlusion. In this situation, combining holistic appearance could gets more enhancement in performance.

As a classical statistical learning method, CCA has been improved since it was proposed in 1936. So, as one of our future work, these improved models can be adopted to improve the proposed pose-robust face recognition approach. Additionally, the accurately aligned facial landmarks play an important role in the proposed method. In real world application it could be an unstable factor. This problem will be studied by introducing elastic matching technique, which has been proved effective for real world face recognition in recent years.

\section*{Acknowledgements}
This work was performed when the first author was a PhD candidate in the Key Lab of Intelligent Information Processing, Institute of Computing Technology, Chinese Academy of Sciences, China. It was supported in part by the Natural Science Foundation of China under Contract 61173065, 61025010, 61222211, and 61003103; and the Beijing Natural Science Foundation (New Technologies and Methods in Intelligent Video Surveillance for Public Security) under Contract 4111003. This work is also partially supported by Singapore Ministry of Education under research Grant MOE2010-T2-1-087.
\bibliographystyle{elsarticle-num}
\bibliography{CCAbib}







\end{document}